\documentclass[letterpaper]{article} 
\usepackage{aaai2026}  
\usepackage{times}  
\usepackage{helvet}  
\usepackage{courier}  
\usepackage[hyphens]{url}  
\usepackage{graphicx} 
\usepackage{natbib}  
\usepackage{caption} 
\usepackage{algorithm}
\usepackage{algorithmic}

\usepackage{amsmath}
\usepackage{amssymb}
\usepackage{multirow}
\usepackage{booktabs}
\usepackage{pifont}

\usepackage{newfloat}
\usepackage{listings}
\DeclareCaptionStyle{ruled}{labelfont=normalfont,labelsep=colon,strut=off} 
\floatstyle{ruled}
\newfloat{listing}{tb}{lst}{}
\floatname{listing}{Listing}
\title{Agent Journey Beyond RGB: Hierarchical Semantic-Spatial Representation Enrichment for Vision-and-Language Navigation}
\author{
    Xuesong Zhang\textsuperscript{\rm 1}\equalcontrib, 
    Yunbo Xu\textsuperscript{\rm 1}\equalcontrib, 
    Jia Li\textsuperscript{\rm 1}\thanks{Corresponding author.}, 
    Ruonan Liu\textsuperscript{\rm 2}, 
    Zhenzhen Hu\textsuperscript{\rm 1}
}
\affiliations{
    \textsuperscript{\rm 1}Hefei University of Technology, Hefei, Chnia \\ \textsuperscript{\rm 2}Shanghai Jiao Tong
University, Shanghai, China \\

    \{xszhang\_hfut, xuyunbocn\}@mail.hfut.edu.cn, jiali@hfut.edu.cn, \\ ruonan.liu@sjtu.edu.cn, huzhen.ice@gmail.com
}

\begin{document}

\maketitle

\begin{abstract}

Navigating unseen environments based on natural language instructions remains difficult for egocentric agents in Vision-and-Language Navigation (VLN). 
Intuitively, humans inherently ground concrete semantic knowledge within spatial layouts during indoor navigation. 
Although previous studies have introduced diverse environmental representations to enhance reasoning, other co-occurrence modalities are often naively concatenated with RGB features, resulting in suboptimal utilization of each modality's distinct contribution. 
Inspired by this, we propose a hierarchical Semantic Understanding and Spatial Awareness (SUSA) architecture to enable agents to perceive and ground environments at diverse scales. 
Specifically, the Textual Semantic Understanding (TSU) module supports local action prediction by generating view-level descriptions, thereby capturing fine-grained environmental semantics and narrowing the modality gap between instructions and environments. 
Complementarily, the Depth-enhanced Spatial Perception (DSP) module incrementally constructs a trajectory-level depth exploration map, providing the agent with a coarse-grained comprehension of the global spatial layout.
Extensive experiments demonstrate that SUSA’s hierarchical representation enrichment not only boosts the navigation performance of the baseline on discrete VLN benchmarks (REVERIE, R2R, and SOON), but also exhibits superior generalization to the continuous R2R-CE.

\end{abstract}

\begin{links}
    \link{Code}{https://github.com/HCI-LMC/VLN-SUSA}
\end{links}


\section{Introduction}

With the advancement of multimodal technology \cite{radford2021clip, li2023blip,zhang2024exploring} in the artificial intelligence community,  embodied VLN tasks \cite{krantz_vlnce_2020,qi2020reverie, EQA_2018_CVPR} have garnered significant attention due to their promising applications in robotics and intelligent assistance. Conventional VLN tasks \cite{anderson2018R2R, krantz_vlnce_2020} focused on step-by-step navigation in unseen environments based on detailed instructions, while goal-oriented VLN tasks \cite{zhu2021soon, qi2020reverie} demand agents to identify predefined objects based on high-level instructions.  Consequently, VLN agents are required to profoundly cognize the environmental spatial information and ground them with textual instructions to brilliantly accomplish navigation tasks.

\begin{figure}
    \centering
    \includegraphics[width=0.95\linewidth]{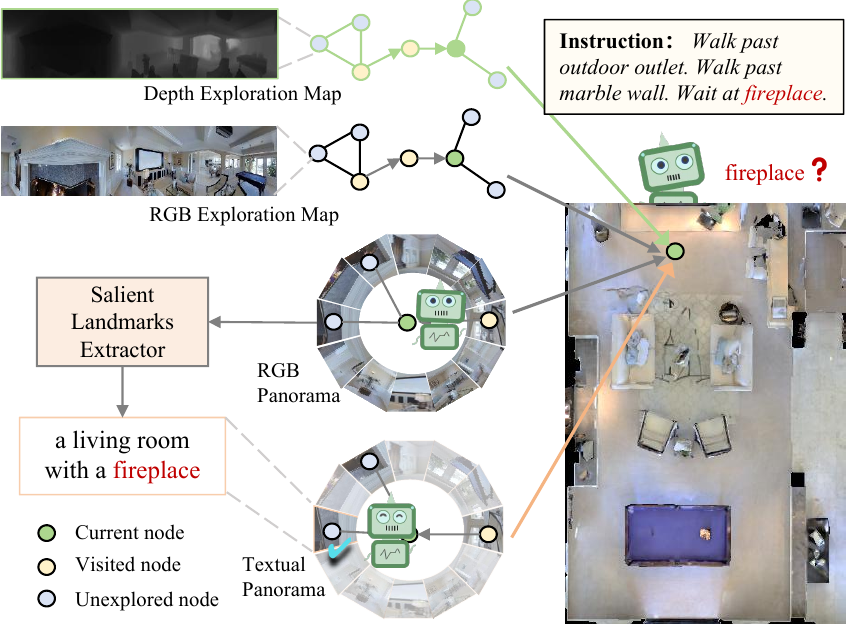}
    \caption{An overview of the proposed SUSA. Beyond RGB inputs, we introduce view-level \emph{textual panoramas} and trajectory-level \emph{depth exploration maps}, supporting the agent’s explicit understanding of the environment.}
    \label{fig:intro}
\end{figure}

Recently, increasing the diversity and quantity of augmented environments \cite{wang2023scaling,wang2023gridmm,liu2024volumetric} has emerged as a popular strategy to facilitate the generalization ability of VLN agents in unseen scenarios. This trend is largely motivated by the limited navigational environments in simulators, as well as the powerful fitting capacity of Transformer-based VLN models~\cite{chen2021hamt, pashevich2021episodic}. 
Nonetheless, VLN agents still struggle to accurately ground the landmarks mentioned in language instructions with the RGB-based visual scene, due to the intrinsic modality heterogeneity.
To address this, recent studies introduce textual semantics as perceptual representations to guide navigation~\cite{pan2024langnav}. 
These approaches leverage candidate object concepts~\cite{wang2023dsrg, lin2023actional}, large language models (LLMs)~\cite{lin2024correctable}, or external knowledge~\cite{li2023kerm} to explicitly reason about textual descriptions of landmarks and enhance environmental grounding. However, these methods tend to combine textual and visual representations in a straightforward manner, which limits the effective utilization of each modality and constrains the interpretability of their individual contributions.
Similarly, co-occurring environmental cues from other modalities—such as frequency~\cite{he2024fda}, spectrum~\cite{hwang2023meta}, and depth~\cite{tan2022depth}—have also been preliminarily explored, though remain underutilized.

In this work, we propose the hierarchical Semantic Understanding and Spatial Awareness (SUSA) architecture, which augments the agent’s environmental comprehension by effectively perceiving and exploiting fine-grained textual semantics and coarse-grained depth-aware spatial information.
To enable fine-grained semantic grounding, as shown in Fig.~\ref{fig:intro}, we construct a view-level textual panorama by extracting salient landmarks as textual semantics, thereby bridging the gap between visual and linguistic modalities. The most instruction-relevant view is then selected for local navigation prediction.
For global action prediction, we progressively construct a trajectory-level depth exploration map to strengthen the agent’s coarse-grained spatial perception.
Finally, SUSA hierarchically aggregates these hybrid semantic-spatial representations to perceive complementary environmental information and employs contrastive learning to align them with natural language instructions. 
Extensive experiments on four VLN benchmarks (R2R~\cite{anderson2018R2R}, REVERIE~\cite{qi2020reverie}, SOON~\cite{zhu2021soon} and R2R-CE~\cite{krantz_vlnce_2020}) demonstrated the superiority of hybrid semantic-spatial representations enrichment.
The contributions of our SUSA are as follows:
\begin{itemize} 
\item SUSA devises a textual-aware semantic understanding (TSU) module to perceive fine-grained textual semantics of environments, and explicitly address modality heterogeneity between instructions and environments.
\item SUSA further constructs a depth exploration map within the depth-enhanced spatial perception (DSP) module to complementarily perceive global environmental layouts and reinforce holistic spatial awareness beyond RGB.
\item Experiments demonstrate that SUSA substantially improves upon the baseline, achieving state-of-the-art performance on three discrete VLN tasks, and exhibits promising generalization in continuous environments.
\end{itemize}


\section{Related Work}

\noindent \textbf{Vision-and-Language Navigation (VLN).}
Vision-and-Language Navigation (VLN) tasks~\cite{anderson2018R2R,qi2020reverie,zhu2021soon}  require an agent to navigate in embodied environments sequentially towards a designated location or object based on natural language instructions. Given that the VLN task can be formulated as a partially observable Markov decision process,
early methods \cite{wang2019reinforced, fried2018speaker} primarily relied on reinforcement learning and imitation learning paradigms to improve navigation performance, incorporating various strategies and auxiliary loss \cite{ wang2019reinforced} designs. More recently, to better model the relationship between the environment and instructions, many researchers \cite{chen2021hamt, pashevich2021episodic}  have focused on transformer-based architectures, which effectively model interactions among instructions, environmental features, and historical navigation trajectories.

\noindent
\textbf{Environment Representations in VLN.}
Limited training sources and environmental representations in VLN tasks hinder the agent to generalize the unknown layouts. To alleviate this limitation, recent endeavors  \cite{wang2023scaling, chen2022learning} augment the quantity of environmental scenes, while others diversify environmental representations by constructing graphs \cite{georgakis2022cross, huang2023visual}, grids \cite{wang2023gridmm, an2023bevbert}, or volumetric environments \cite{liu2024volumetric}. 
Beyond RGB inputs, recent efforts have explored other co-occurrence multimodal\cite{li2023kerm,he2024fda} cues to enrich environmental representations. 
For instance, SEAT \cite{wang2024enhanced} enhances cross-domain environmental perception by querying the Matterport3D simulator~\cite{Matterport3D} for depth and semantics. However, SEAT fuses these multi-type representations into a unified space without explicitly grounding each modality. Moreover, these semantic representations \cite{wang2024enhanced,hong2023learning} are not inherently textual, limiting their ability to bridge the modality gap. Other approaches incorporate depth information~\cite{an2023bevbert, wang2023gridmm}; however, they typically project depth values onto the spatial coordinates of RGB images without independently grounding the depth modality. 
In contrast, our method constructs the view-level textual panoramas and trajectory-level depth exploration maps, individually grounding them with instructions to perceive rich environmental representations.

\begin{figure*}[ht!]
    \centering
    \includegraphics[width=0.95\linewidth]{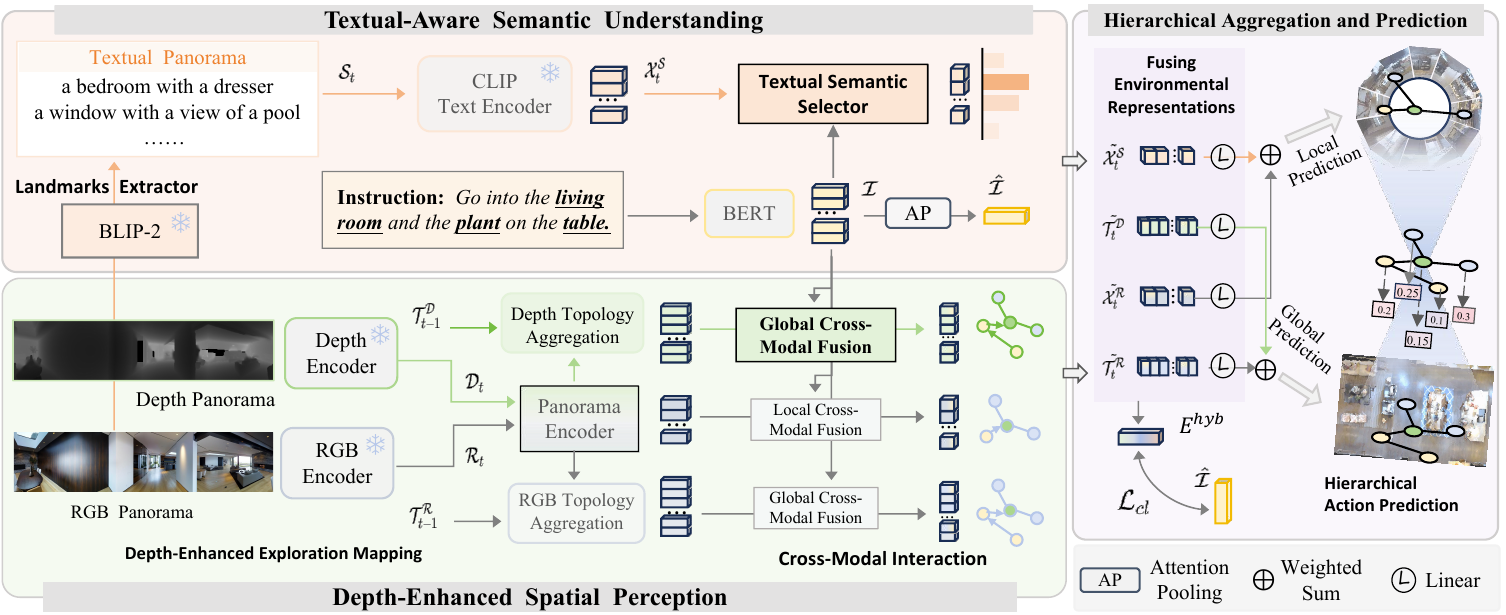}
    \caption{The detailed architecture of our proposed SUSA model. The orange and green arrows highlight our proposed Textual-Aware Semantic Understanding (\emph{TSU}) and Depth-Aware Spatial Perception (\emph{DSP}) modules, respectively. The Hierarchical Aggregation and Prediction (\emph{HAP}) module is designed for hierarchical aligning environmental representations with instructions.}
    \label{fig:SUSA}
\end{figure*}

\section{Methodology}

\subsection{Preliminary}
\noindent \textbf{Task Formulation.}
We focus on discrete VLN tasks \cite{anderson2018R2R,qi2020reverie,zhu2021soon} within a Matterport3D simulator \cite{Matterport3D}, modeled as an undirected graph $\mathcal{G} = \{\mathcal{N}, \mathcal{E}\}$, where $\mathcal{N}$ denotes nodes and $\mathcal{E}$ signifies the connectivity paths. At the outset of each episode, the agent equipped with a GPS sensor, RGB and depth cameras, is situated at a randomly navigable node and receives a language instruction $I = \{\text{w}_i\}_{i=1}^l$ including $l$ words. We use BERT \cite{devlin2019bert} to extract instruction features and then embed them as $\mathcal{I} = \{w_1, \dotso, w_l\}$.
At each time step $t$, the agent can perceive the surrounding RGB panorama $R_t = \{\text{r}_{t,i}\}_{i=1}^{n=36}$ and depth panorama $D_t = \{\text{d}_{t,i}\}_{i=1}^{n=36}$ along with $n$ corresponding views. 
During navigation, the agent makes the next action $a_t$ by selecting one navigable node from the candidate actions $\mathcal{A}_t = \{a_{t,i}\}_{i=0}^{k}$. 

\noindent \textbf{Overview of the Proposed Method.}
Our objective is to enrich complementary environmental content by equipping the agent with a textual semantic panorama and a depth exploration map.  Following prevailing approaches \cite{he2024fda}, we adopt DUET \cite{chen2022duet} as our baseline, which merely utilizes RGB panoramas and topological maps as environmental representations for both local and global action predictions.  The SUSA framework consists of three main modules: (a) \textit{TSU}: textual-aware semantic selection is detailed in Section~\ref{TSA}, (b) \textit{DSP}: depth-enhanced visual environment exploration is presented in Section~\ref{DVE}, and (c) \textit{HAP}: hybrid representation aggregation and hierarchical action prediction is subsequently introduced in Section~\ref{HFP}, as illustrated in Fig.~\ref{fig:SUSA}.

\subsection{Textual-Aware Semantic Understanding}
\label{TSA}

In this module, we build a textual semantic panorama that enables the agent to understand the environment from a textual perspective, allowing it to select the most relevant navigable view as the optimal local navigation action.
\begin{figure}[t]
    \centering
    \includegraphics[width=0.9\linewidth]{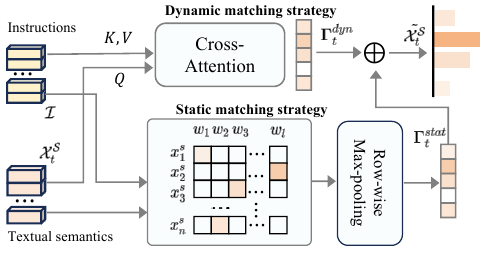}
    \caption{Illustration of the \emph{TSU} module in Fig.\ref{fig:SUSA}, which matches instructions and semantic features by static and dynamic matching strategies.}
    \label{fig:pip-sem}
\end{figure}

\noindent \textbf{Salient Landmarks Extractor.}
The pretrained visual-language model readily provides the capability for agents to perceive textual semantics by generating detailed captions from RGB images. As illustrated in Fig.~\ref{fig:SUSA}, we utilize off-the-shelf  BLIP-2-FlanT5-xxL \cite{li2023blip}, a powerful visual-language model, to discretize $R_t$ and generate corresponding textual descriptions for each view using the simple prompt ``a photo of". Following \cite{gao2024enhancing,li2023panogen}, this process yields a textual panorama \( \mathcal{S}_t = \{s_{t,i}\}_{i=1}^{n=36} \), where each view $s_{t,i}$ contains informed descriptions of salient landmarks. These descriptions typically involve room types and objects (e.g.,``a living room with two chairs"), which assist in the object recognition. We further discuss the quality of the generated descriptions in supplementary materials. Additionally, we employ the CLIP (ViT-L-14-336px) text encoder \cite{radford2021clip} to extract textual semantic panorama features,  embedding them as \( \mathcal{X}^\mathcal{S}_t  = \{x^s_{t,1}, \dots, x^s_{t,n}\} \). 
Thanks to the powerful yet frozen BLIP-2 and CLIP models, we empower the agent to obtain textual semantic features from the navigable nodes in its vicinity during navigation.

\noindent \textbf{Textual Semantic Selector.}
Similar to \cite{pan2024langnav}, we treat the textual semantic as a perceptual representation for navigation in this module,  directly matching salient landmarks with the instructions.
As shown in Fig.~\ref{fig:SUSA} and Fig.~\ref{fig:pip-sem}, our textual semantic selector combines both $static$ and $dynamic$ matching strategies to select the most instruction-relevant view in the textual panorama $\mathcal{X}^\mathcal{S}_t$.

\textit{Static Matching:} The most straightforward approach to semantic alignment is by calculating similarity, where higher similarity indicates stronger relevance between the textual semantics of each view and the instruction.
Consequently, we calculate the cosine similarity matrix \(\mathbf{M} \in \mathbb{R}^{n \times l}\), which represents the correlation between each word in the instruction \(\mathcal{I}\) and each view in the textual panorama \(\mathcal{X}^\mathcal{S}_t\). 
Nevertheless, the textual semantics generated only correspond to salient landmarks (e.g.,``plant," or ``table") rather than actions (e.g., ``go to" or `` turn left") in the instructions. Therefore, we then apply row-wise max-pooling to purify the static instruction relevance \(\Gamma_t^{stat} \in \mathbb{R}^{n}\) to match the most relevant landmarks between textual semantics and instructions at step \(t\) :
\begin{equation}
\small
\begin{split}
    \mathbf{M}_{i,j} = \frac{\mathcal{X}^\mathcal{S}_{t,i} \cdot \mathcal{I}_j}{\|\mathcal{X}^\mathcal{S}_{t,i}\| \|\mathcal{I}_j\|}, \ \ \ \ \Gamma_t^{stat} = \max_{j} \mathbf{M}_{i,j}
\end{split}
    \label{equation: DynamicFuse}
\end{equation}
where \(i = 1, \dots, n\) and \(j = 1, \dots, l\). Although straightforward, the static strategy relies on the monotonic similarity matrix to match textual semantics \(\mathcal{X}^\mathcal{S}_t\) and instructions \(\mathcal{I}\), struggling with long-term navigation reasoning. 

\textit{Dynamic Matching:} Furthermore, a dynamic matching strategy is introduced to better capture the intricate relationships between the environment and instruction. Specifically, we utilize a standard multi-layer transformer decoder to dynamically match the relevance $\Gamma_t^{dyn}$ between textual semantics and instructions, as formalized by:
\begin{equation}
\small
\begin{split}
\Gamma_t^{dyn}  =  \text{TCA} (\mathcal{X}^\mathcal{S}_{t}, \mathcal{I}, \mathcal{I}),
\end{split}
    \label{equation: SAM}
\end{equation}
where TCA denotes a \textbf{t}extual modality \textbf{c}ross-\textbf{a}ttention mechanism that models textual semantic-instruction relationships. TCA comprises a cross-attention layer, a feed-forward network (FFN) layer, and LayerNorm, enabling the agent to adapt to long-term navigation sequences.


To trade off computational efficiency and navigation accuracy, we flexibly combine the relevance of static  \(\Gamma_t^{sim}\) and dynamic \(\Gamma_t^{att}\).  The latent semantic panoramic feature \(\tilde{\mathcal{X}_t^S}\) is expressed as follows:
\begin{equation}
\begin{split}
    \tilde{\mathcal{X}_t^S} =  \delta \ast \Gamma_t^{stat} + (1 - \delta) \ast \Gamma_t^{dyn},
\end{split}
    \label{equation: text}
\end{equation} where the balance factor \(\delta\) regulates the relative contributions of each matching strategy. By mapping the environment to the same textual modality, the textual semantic selector enables the agent to explicitly understand the environment context from a textual semantic perspective, thus facilitating grounded language instructions.

\subsection{Depth-Enhanced Spatial Perception}
\label{DVE}
Depth images, which also contain more intuitive and easily distinguishable structural information, were not adequately exploited in earlier works \cite{tan2022depth}. To enhance the spatial perception, we concurrently construct a depth-based exploration map to memorize depth-based navigation trajectory alongside the RGB-based one \cite{chen2022duet}.


\noindent \textbf{Depth-Enhanced Exploration Mapping.}
Concretely, we obtain ground truth depth images for each view from the Matterport3D simulator.
As illustrated in Fig.~\ref{fig:SUSA}, we then employ ResNet-50 \cite{he2016resnet} (pretrained on the Gibson \cite{xia2018gibson}) and CLIP (ViT-L/14@336px) as the depth and RGB encoders, respectively, to extract panoramic depth features $\mathcal{D}_t \in \mathbb{R}^{n\times d_v}$ and panoramic RGB features $\mathcal{R}_t \in \mathbb{R}^{n\times d_v}$, where $d_v$ represents the dimension of each view.
Moreover, we employ a two-layer transformer-based panorama encoding module to perform self-attention and sequentially encode the depth and RGB panoramic images. The panorama encoding modules for both modalities share weights, since each view in the depth and RGB panoramas is accordant.
Subsequently, we aggregate each visited node and its adjacent views by average pooling to construct depth and RGB exploration maps in parallel. 
The explored depth and RGB trajectory maps are formalized as $\mathcal{T}_{t}^\mathcal{D} \in \mathbb{R}^{k \times d_n}$ and $\mathcal{T}^\mathcal{R}_t \in \mathbb{R}^{k \times d_n}_t$ . Here, $k$ denotes the number of navigable nodes at step $t$, while $d_n$ represents the dimension of each navigable node. These exploration maps memorize the agent navigation trajectory information for global action predictions, which serve to guide the agent traversal across various navigable nodes when considering backtracking.

\noindent \textbf{Cross-Modal Interaction and Reasoning.}
To defer to instructions, we utilize a multi-layer cross-encoder that interacts with both the language instructions and the depth exploration map features to generate the refined depth map  $\tilde{\mathcal{T}^\mathcal{D}}$. This encoder comprises four residual connection layers of a cross-modal transformer, each incorporating graph-aware self-attention (GASA)~\cite{chen2022duet}, cross-modal attention, and a feed-forward network.
The GASA operation is formally expressed as $\text{GASA}(X) = \eth(X\Theta_q (X\Theta_k)^{T} + \mathcal{M}) X \Theta_v$
where $\mathcal{M}$ represents a learnable distance matrix encoding the relative distances between navigation nodes, $\eth$ denotes the Softmax activation function, and $\Theta_q, \Theta_k, \Theta_v$ are learnable weight matrices.
In parallel, we facilitate similar cross-modal interactions between the language instructions and the RGB exploration map $\mathcal{T}^\mathcal{R}$, as well as the RGB panoramas $\mathcal{X}^\mathcal{R}$. These are encoded as $\tilde{\mathcal{T}^\mathcal{R}}$ and $\tilde{\mathcal{X}^\mathcal{R}}$, thereby enabling the model to perform both global and local predictions based on the given instructions $\mathcal{I}$.
Compared to the standalone $\mathcal{T}^\mathcal{R}$, the incorporation of a depth exploration map $\mathcal{T}^\mathcal{D}$ enhances the perceptual clarity of environmental structures, mitigating the risk of overfitting to RGB-based visual noises \cite{ilinykh2022look-biasvln} or biases \cite{hu2019areyou-biasvln,wang2024vision}.



\begin{figure}[t]
    \centering
    \includegraphics[width=0.95\linewidth]{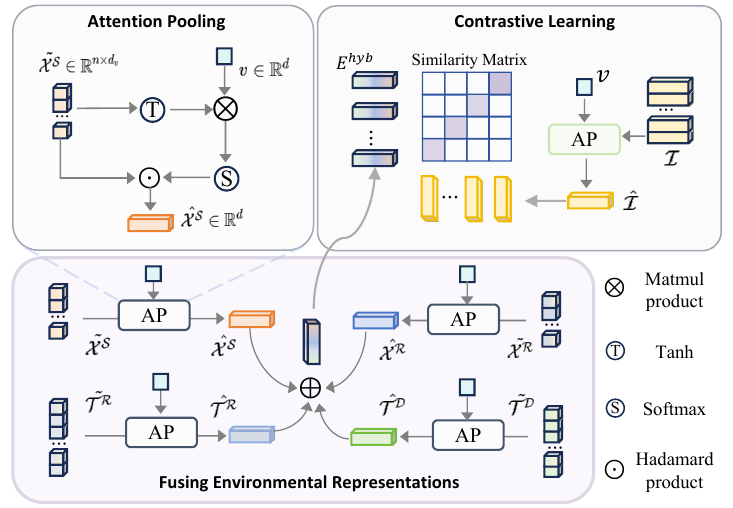}
    \caption{The pipeline for fusing hybrid environmental representations in the proposed \emph{HAP} module in Fig.~\ref{fig:SUSA}.}
    \label{fig:4}
\end{figure}
\subsection{Hierarchical Aggregation and Prediction} \label{HFP} 
Building on aforementioned methods, at the current state $t$, the agent synchronously perceives the environment through multiple modalities—textual semantic, depth, and RGB —at various scales (local and global). To encourage the agent to aggregate complementary context from these hybrid representations, we encode these features into a low-dimensional latent space via attention pooling and apply contrastive learning to align them with the instructions.

\noindent \textbf{Hybrid Representations Aggregation.}
As depicted in Fig.~\ref{fig:4}, we employ attention pooling to project all environmental representations at step $t$, along with the corresponding instruction, into an equivalent dimension $d$. Specifically, the attention pooling $\text{AP}$ operation is defined as follows:
\begin{equation}
    \hat{\mathcal{H}} = \text{AP}(\mathcal{H}, v), \quad \mathcal{H} \in \{\tilde{\mathcal{X}^\mathcal{S}}, \tilde{\mathcal{T}^\mathcal{D}}, \tilde{\mathcal{X}^R}, \tilde{\mathcal{T}^R}, \mathcal{I}\},
\end{equation}
where $v \in \mathbb{R}^{d}$ denoting the learnable query vector and $\hat{\mathcal{H}} \in \mathbb{R}^d$. The $\text{AP}$ operator is given by:
\begin{equation}
    \eta = \eth\left(\tau(\mathcal{H}) \otimes v^\top\right), \quad \hat{\mathcal{H}} = \tau\left(\eta \odot \mathcal{H}\right),
\end{equation}
with $\tau$ representing the tanh activation function.
Subsequently, we perform weighted aggregation of the hybrid environmental representations $E^\mathrm{hyb}$, as follows:
\begin{equation}
    \small
    \mathcal{B} = \begin{bmatrix}\beta_1 &\beta_2 & \beta_3 & \beta_4 \end{bmatrix}^\top =  \sigma ( \mathrm{FFN} [\tilde{\mathcal{X}_0^\mathcal{S}} ;  \tilde{\mathcal{X}_0^\mathcal{R}}; \tilde{\mathcal{T}_0^\mathcal{D}} ; \tilde{\mathcal{T}_0^\mathcal{R}}  ] ) 
    \label{eq: branch-weights}
\end{equation}
\begin{equation}
\small
    E^{\mathrm{hyb}} =  \begin{bmatrix} \hat{\mathcal{X}^\mathcal{S}}&\hat{\mathcal{X}^\mathcal{R}}&\hat{\mathcal{T}^\mathcal{D}}&\hat{\mathcal{T}^\mathcal{R}} \end{bmatrix} \cdot \mathcal{B},
\end{equation}
where $\sigma$ stands for the sigmoid activation function, and $[;]$ denotes concatenation.

\noindent \textbf{Hierarchical Action Prediction.}
We perform hierarchical action prediction by synergistically fusing the prediction scores from multiple branches to capture complementary information.  Specifically, the scores for each branch are computed via separate $\mathrm{FFN}$: $P_\mathcal{\bar{H}} = \mathrm{FFN}_\mathcal{\bar{H}}(\mathcal{\bar{H}})$, where $\mathcal{\bar{H}} = \{ x \in \mathcal{H} \mid x \neq \mathcal{I} \}$.
Then, we apply weighted fusion to the textual semantic panorama $\mathcal{X}^\mathcal{S}$ and the RGB panorama $\mathcal{X}^\mathcal{R}$ to make local predictions, along with global predictions derived from exploration maps based on depth $\mathcal{T}^\mathcal{D}$ and RGB  $\mathcal{T}^\mathcal{R}$, which is formulated as:
\begin{gather}
    P_\mathcal{X} = \beta_1 P_{\mathcal{X}^\mathcal{S}} + \beta_2 P_{\mathcal{X}^\mathcal{R}}, \\
    P_\mathcal{T} = \beta_3 P_{\mathcal{T}^\mathcal{D}} + \beta_4 P_{\mathcal{T}^\mathcal{R}}.
\end{gather}
We convert local predictions into global actions and predict $a_t$ via dynamic fusing~\cite{chen2022duet} $P_\mathcal{T}$ and $P_\mathcal{X}$.

\subsection{Training}
\label{fusion}

\noindent \textbf{Partially Pretraining.}
Our pre-training phase followed prior work~\cite{chen2022duet}, utilizing auxiliary tasks such as Masked Language Modeling (MLM), Masked Region Classification (MRC), and Single-step Action Prediction (SAP). Object Grounding (OG) task to localize target objects in REVERIE and SOON.
Given the sparsity of depth images and textual semantics compared to RGB images, completely pre-training all parameter of SUSA may lead to overfitting and increased training costs. As shown in Fig.~\ref{fig:SUSA}, only the computation flows indicated by black arrows were pre-trained. The effectiveness of this partial pre-trained strategy is demonstrated in Section~\ref{ablation}.

\begin{table*}[ht]
\small
\centering
\setlength{\tabcolsep}{6pt}
\renewcommand{\arraystretch}{1.0}
\begin{tabular}{l|ccc|ccc|ccc|ccc}
\toprule
\multirow{3}{*}{\textbf{Methods}} & 
\multicolumn{6}{c}{\textit{\textbf{REVERIE}}} & 
\multicolumn{6}{c}{\textit{\textbf{R2R}}} \\
\cmidrule(lr){2-7} \cmidrule(lr){8-13}
 & \multicolumn{3}{c|}{\textbf{Val Unseen}} & 
 \multicolumn{3}{c|}{\textbf{Test Unseen}} & 
 \multicolumn{3}{c|}{\textbf{Val Unseen}} & 
 \multicolumn{3}{c}{\textbf{Test Unseen}} \\
& SR & \textbf{SPL} & RGSPL & 
 SR & \textbf{SPL} & RGSPL & 
 SR & \textbf{SPL} & NE$\downarrow$ & 
 SR & \textbf{SPL} & NE$\downarrow$ \\
\midrule
RCM \cite{wang2019reinforced}  & 9.2& 6.9  & 3.8  & 7.8 & 6.6  & 3.1 & 43 & -  &6.0  &43  &38    &6.1  \\
GridMM  \cite{wang2023gridmm}   & 51.3 & 36.4 & 24.5 & 53.1 & 36.6 & 23.4 & \textbf{75} & 64 & -  &\textbf{73}  &62  & -   \ \\
KERM \cite{li2023kerm} & 49.0 & 34.8 & 24.1 & 52.2 & 37.4 & 23.1 &71.9  & 60.9 &3.2  &69.7  &59.2  &3.6  \  \\
AZHP  \cite{zhan2024enhancing} & 49.0 & 36.2 & 24.1 & 52.5 & 36.1 & 22.5 &71  &60  &3.2  &69  &59  &3.4  \ \\
FDA  \cite{he2024fda}   & 47.5 & 35.9 & 24.3 & 49.6 & 36.4 & 22.0 &72  &64  &3.4  &69  &62  &3.41  \  \\
CONSOLE \cite{lin2024correctable}   & 50.0 & 34.4 & 23.3 & \textbf{55.1} & 37.1 & 22.2 &73  &63  &3.0  &72  &61  &3.3  \ \\
ACME \cite{wu2025adaptive} & 49.5 & 32.4 & 24.0 & 51.9 &34.7 & 23.6 &72.8  & 62.3 &3.12  &70.4  &61.2  &3.68  \  \\
\midrule
DUET \cite{chen2022duet}  & 46.9 & 33.7 & 23.0 & 52.5 & 36.0 & 22.0 &72  &60  &3.3  &69  &59  &3.65  \ \\
\textbf{SUSA (Ours)} & \textbf{51.7} & \textbf{38.8} & \textbf{26.5} & 54.3 & \textbf{41.5} & \textbf{27.3} &73.0  &\textbf{64.8}  &\textbf{3.0}  &72.5  &\textbf{63.8}  &\textbf{3.20}  \ \\
\bottomrule
\end{tabular}
\caption{Comparison with the state of the art on REVERIE and R2R. \textbf{Bold} highlights the best performance in each column. $\downarrow$ indicates better performance with lower values. ``–'': unavailable statistics. \% is omitted for brevity.}
\label{tab:reverie-sota}
\end{table*}

\noindent \textbf{Contrastive and Imitation Learning.} 
During fine-tuning, we employ a contrastive learning loss $\mathcal{L}_{cl}$ to align the hybrid environmental representations $E^{hyb}$ with the corresponding instructions $\mathcal{I}$:
\begin{equation}
    \begin{aligned}
        \textstyle
    \mathcal{L}_{cl} = - \frac{1}{2B} \sum_{i=1}^{B} \log \frac{\exp((E^\mathrm{hyb}_{i}, \mathcal{I}_{i}) / t)}{\sum_{j=1}^{B} \exp((E^\mathrm{hyb}_{i}, \mathcal{I}_{j}) / t)} \\
    \textstyle
- \frac{1}{2B} \sum_{i=1}^{B} \log \frac{\exp((\mathcal{I}_{i}, E^\mathrm{hyb}_{i}) / t)}{\sum_{j=1}^{B} \exp((\mathcal{I}_{i}, E^\mathrm{hyb}_{j}) / t)},
    \end{aligned}
\end{equation} where $B$ and $t$ represent the batch size and the temperature. 
Then, we train the model based on hybrid environment representations by performing single-step action prediction \cite{chen2021hamt, chen2022duet} with a contrastive learning loss $\mathcal{L}_{cl}$. 
The hybrid single action prediction loss in imitation learning is computed as $\mathcal{L}_{hsap}^{gt} = \sum_{t=1}^{T}  -\log p(a_t^{gt} | \mathcal{I}, \mathcal{S}_t, \mathcal{D}_{<t},\mathcal{R}_{<t})$ and $\mathcal{L}_{hsap}^{\star} = \sum_{t=1}^{T} - \log p(a_t^{\star} | \mathcal{I}, \mathcal{S}_t, \mathcal{D}_{<t},\mathcal{R}_{<t})$, where $a_t^{gt}$ represents the ground truth action, and $a_t^{\star}$ denotes the pseudo-target action sampled from current exploration map. The final training loss is then expressed as follows:
\begin{equation}
    \mathcal{L}_{SUSA} =  \lambda_1 \mathcal{L}_{hsap}^{gt} + \mathcal{L}_{hsap}^{\star} + \lambda_2 \mathcal{L}_{cl}
\label{loss}
\end{equation}
where $\lambda_1$ and $\lambda_2$ denote loss weights.

\begin{table}[t]
\centering
\setlength\tabcolsep{2pt}
\begin{tabular}{@{}l|cc|cc@{}}
\toprule
\multirow{2}{*}{\textbf{Method}} & \multicolumn{2}{c}{\textbf{Val Unseen}} & \multicolumn{2}{c}{\textbf{Test Unseen}} \\
  & \textbf{SPL} & RGSPL  & \textbf{SPL} & RGSPL\\ \midrule
GridMM~\cite{wang2023gridmm}   & 24.8 & 3.9  & 21.2 & 4.1  \\
KERM~\cite{li2023kerm} &23.1&4.4 &- & -\\
SEAT~\cite{wang2024enhanced} &24.8&3.9  &22.5 & 4.4\\
ACME~\cite{wu2025adaptive} &27.8 &4.3 &22.3 & 5.5\\
\midrule
DUET~\cite{chen2022duet}    & 22.5 & 3.7  & 21.4 & 4.1  \\
\textbf{SUSA(Ours)} & \textbf{30.8} &\textbf{6.4}  & \textbf{25.4} & \textbf{5.9}   \\     \bottomrule         
\end{tabular}
\caption{Comparison with the state of the art on the SOON.}
\label{tab:soon-sota}
\end{table}

\begin{table}[t]
\centering
\setlength\tabcolsep{2pt}
\begin{tabular}{@{}l|cc|cc@{}}
\toprule
\multirow{2}{*}{\textbf{Methods}} & 
\multicolumn{2}{c|}{\textbf{Val Unseen}} & 
\multicolumn{2}{c}{\textbf{Test Unseen}} \\
& SR & \textbf{SPL}  & SR & \textbf{SPL}  \\
\midrule
GridMM~\cite{wang2023gridmm} & 49 & 41  & 46 & 39  \\
AO-Planner~\cite{chen2025affordances}& 47& 33& -& -\\
ETPNav~\cite{an2024etpnav} & 57 & 49  & 55 & 48  \\
\midrule
DUET~\cite{chen2022duet} & 51.9 & 41.2 & 47.1 & 40.5  \\
\textbf{SUSA (Ours)}  & \underline{52.7} & \underline{43.6} & \underline{50.9} & \underline{43.9}  \\
\bottomrule
\end{tabular}
\caption{Comparison with the state of the art on R2R-CE. We directly transfer the DUET and SUSA as planners of ETPNav into the continuous environment without pretraining. \underline{Underline} indicates improvements over the baseline.}
\label{tab:r2r-ce}
\end{table}

\section{Experiments}

\subsection{Task Setup and Implementation}

\noindent \textbf{Datasets.}
We mainly evaluate our model on three diverse VLN benchmarks.
\textbf{R2R} \cite{anderson2018R2R} focuses solely on navigation following detailed instructions.
\textbf{REVERIE} \cite{qi2020reverie} requires the agent to recognize the correct object from candidate bounding boxes upon reaching the navigation goal. 
\textbf{SOON} \cite{zhu2021soon} challenges agents to generate candidate bounding boxes using an object detector.
Given that discrete agents can serve as planners of ETPNav \cite{an2024etpnav}, we also evaluate SUSA on the continuous R2R-CE \cite{krantz_vlnce_2020}.
\textbf{R2R-CE} is constructed from the discrete Matterport3D and executed in the continuous Habitat simulator~\cite{savva2019habitat}.

\noindent \textbf{Metrics.}
We evaluate the navigation performance using standard metrics:
Success Rate (SR),
Oracle Success Rate (OSR),
Navigation Error (NE),
and Success Rate weighted by Path Length (SPL).
For the REVERIE and SOON tasks, which involve object identification, we also evaluate object grounding metrics: Remote Grounding Success (RGS) and RGS weighted by Path Length (RGSPL). Given that \textbf{SPL} optimally balances navigation success rate and trajectory length, we employ it as our primary metric. 


\noindent \textbf{Implementation Details.} 
Pre-training was performed for 100k iterations with a batch size of 32. Fine-tuning was performed for 25k iterations for VLN tasks. For fine-tuning, we used batch sizes of 4, 8, 2 and 16 for R2R, REVERIE, SOON and R2R-CE, respectively.
The default hyperparameters are $\lambda_1 $ = 0.2 and $\lambda_2$= 0.8. The node and view dimensions are identical: $d_v = d_n = d = 768$. We minimized changes to the baseline (DUET)~\cite{chen2022duet} settings and used no additional annotations.
All experiments were conducted on a \textbf{single} NVIDIA RTX 4090 GPU.

\subsection{Comparisons with State of the Art}

We primarily conduct a comprehensive comparison on three discrete VLN tasks—REVERIE~\cite{qi2020reverie}, R2R~\cite{anderson2018R2R}, and SOON~\cite{zhu2021soon}.
For fairness, we exclude methods that involve pre-exploration \cite{ sigurdsson2023rrex} or utilize large-scale data augmentation \cite{wang2023scaling}.
Tables~\ref{tab:reverie-sota} and~\ref{tab:soon-sota} present a comparative analysis on REVERIE, R2R, and SOON. Our method achieves, and in some cases surpasses, the performance of previous state-of-the-art approaches on various metrics.
For instance, SUSA achieves an SPL/RGSPL of 41.5\%/27.3\% on the REVERIE test unseen split, significantly outperforming the baseline (DUET) by a large margin of 5.5\%/5.3\%. 

Furthermore, we briefly transfer the agent into continuous environments to coarsely assess its adaptability and generalization.
Results in Table~\ref{tab:r2r-ce} indicate that SUSA achieves superior performance over the baseline in the continuous R2R-CE task, highlighting the potential of hierarchical representation enrichment in continuous navigation scenarios.


\subsection{Diagnostic Experiment}
\label{ablation}

\begin{table}[t]
\centering
\setlength{\tabcolsep}{6pt}
\begin{tabular}{cc|cccc}
\toprule
    \textbf{RGB} & \textbf{Depth}  & SR  & \textbf{SPL}  & RGS  & RGSPL  \\ \midrule
ViT & \textit{$w/o$} & 46.9 & 33.7 & 32.1 & 23.0 \\
CLIP & \textit{$w/o$} & 51.5 & 35.7 & \textbf{35.1} & 24.4 \\
CLIP & ImageNet  & 50.1 & 34.6& 34.3 & 23.6 \\
CLIP & Gibson   & \textbf{52.0} & \textbf{39.2} & 35.0 & \textbf{26.5} \\
\bottomrule
\end{tabular}

\caption{Ablation of different visual representations on the REVERIE validation unseen split. Gibson and ImageNet serve as pre-training sources for the depth encoder (ResNet-50) in the DSP module.}
\label{tab:visual-comparison}
\end{table}


\begin{table}[]
\centering
\setlength{\tabcolsep}{4pt}
\begin{tabular}{c|ccc|ccc}
\toprule
\multirow{2}{*}{$\delta$} & \multicolumn{3}{c}{\textit{\textbf{REVERIE}}} & \multicolumn{3}{c}{\textit{\textbf{R2R}}} \\
 & SR& \textbf{SPL} & RGSPL & SR& \textbf{SPL}& nDTW\\ \midrule
\textit{$w/o$}&52.0 &39.2  &26.5 &72.8 &62.7 &69.3\\
0 & 53.9 &38.8& \textbf{26.8}& 72.2 &63.7 & 70.7 \\
0.5& \textbf{55.0}& \textbf{39.5} &26.7& \textbf{73.0}&\textbf{64.8}&\textbf{71.0}  \\
1.0&53.5&37.8&25.9&72.3&62.9&69.1 \\
$adaptive$ & 52.4&38.2 &26.2 & 71.9& 63.1&69.9 \\ 
\bottomrule
\end{tabular}
\caption{Ablation of static and dynamic matching strategies in the TSU module, balanced by the balance factor $\delta$.}
\label{tab:semantic-module}
\end{table}

\noindent \textbf{1) Image Features vs. Structural Features.}
We analyze RGB and depth features contributions within the DSP module (Table ~\ref{tab:visual-comparison}).
For RGB images, substituting ViT \cite{dosovitskiy2020vit} features with those from CLIP \cite{radford2021clip} yields marked improvements across all metrics.
As for depth features, we leveraged depth features extracted by ResNet-50~\cite{he2016resnet} pre-trained on the ImageNet~\cite{deng2009imagenet} or Gibson~\cite{xia2018gibson} datasets to investigated the effect of depth exploration maps.
Compared to ImageNet, the Gibson simulator provides spatial structural information that serves as valuable navigation priors. Therefore, the ResNet-50 pre-trained on the Gibson 
yields notable improvements in the SPL metric without compromising SR or RGS. These improvements indicate that better spatial perception enhances navigation efficiency.



\noindent \textbf{2) Static Matching vs. Dynamic Matching.}
To assess the impact of static matching ($\delta=0$) and dynamic matching ($\delta=1$) strategies in the TSU module, we perform ablation studies based on balance factor $\delta$ on the R2R and REVERIE.
Table~\ref{tab:semantic-module} summarizes these results, where \textit{w/o} indicates the exclusion of the TSU module.
Furthermore, \textit{adaptive} denotes that $\delta$  is a learnable parameter which adaptively adjusts the contribution of each matching strategy.
While the TSU module can achieve superior performance under the \textit{adaptive} setting, it underperforms when $\delta$ = 0.5. Specifically, at $\delta$ = 0.5, the SPL reaches 39.50\% and 64.85\% on the two datasets. Excellent results demonstrate dual merits of TSU module, which boosts object identification accuracy on the REVERIE while simultaneously improving the fidelity between navigation trajectories and instructions on the R2R.

\begin{figure}
    \centering
    \includegraphics[width=1\linewidth]{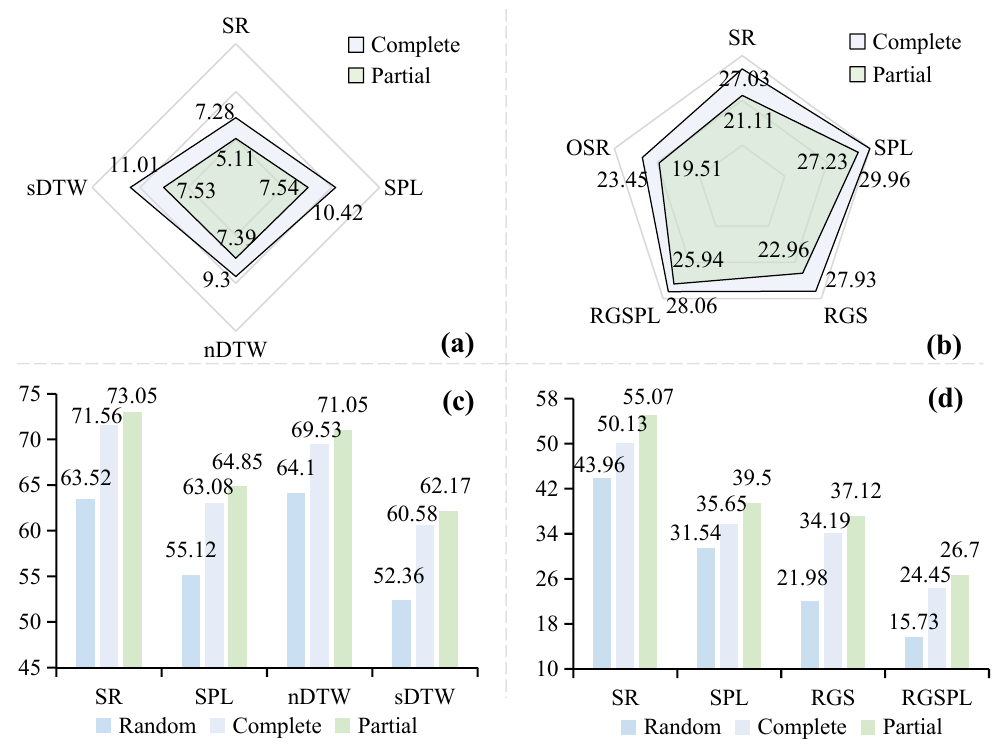}
    \caption{(a) and (b) show the performance gap (\textbf{$\downarrow$}, lower values indicate better performance.) between seen and unseen environments, while (c) and (d) present key metrics (\textbf{$\uparrow$}) for different pretraining strategies on the R2R and REVERIE.}
    \label{fig:init}
\end{figure}

\begin{table}[t]
\centering
\setlength{\tabcolsep}{4pt}
\begin{tabular}{c|ccc|ccc}
\hline
ID & \textbf{DSP} & \textbf{TSU} & \textbf{HAP}& SR& SPL & RGSPL                      \\ \hline
1  & \ding{55} & \ding{55} & \ding{55} &46.9$|$52.5 & 33.7$|$36.0 & 23.0$|$22.0 \\
2  & \ding{51} & \ding{55} & \ding{55} & 52.0$|$52.5 & 39.2$|$39.5 & 26.5$|$24.9 \\
3  & \ding{55} & \ding{51} & \ding{55} & 53.1$|$53.8 & 38.3$|$40.8 & 26.7$|$26.4 \\
4  & \ding{51} & \ding{51} & \ding{55} & 55.0$|$51.9 & 39.5$|$37.8 & 26.7$|$25.2 \\
5  & \ding{51} & \ding{51} & \ding{51} & 51.7$|$54.3 & 38.8$|$41.5 & 26.5$|$27.3 \\ \hline
\end{tabular}
\caption{Ablation study on different components of SUSA on the REVERIE (validation$|$test) unseen split.}
\label{tab:reverie-overall}

\end{table}

\noindent \textbf{3) Improved Generalization via Pretraining.}
As aforementioned in Section~\ref{fusion}, pretraining the entire SUSA framework on multiple auxiliary tasks may lead to overfitting, which negatively impact the generalization in unseen environments.
As shown in Fig.~\ref{fig:init}, the partial pretraining strategy not only consistently narrows the performance gap between seen and unseen environments but also outperforms complete/random pretraining strategies in navigation performance. In Fig. \ref{fig:init} (a), the partial pretraining model achieves an SR gap of only 21.11\% between the seen and unseen splits on REVERIE, reducing the gap by 5.92\% compared to the complete pretraining strategy. Additionally, as depicted in Fig.~\ref{fig:init} (c), multiple metrics indicate that partial pretraining results in superior performance. Fig.~\ref{fig:init} (b) and (d) illustrate similar trends are observed on R2R, further emphasizing the generalization of the partial pretraining strategy.



\noindent  \textbf{4) Overall Design.}
To thoroughly evaluate the efficacy of key components, we conduct ablation experiments on the REVERIE validation and test unseen splits in Table~\ref{tab:reverie-overall}. 
An intriguing phenomenon is observed in Table~\ref{tab:reverie-overall}: directly incorporating the DSP and TSU modules (\#4) results in an improvement on validation unseen (e.g., SR=55.07, SPL=39.50), while performance on test unseen deteriorates. 
Conversely, contrastive learning within the HAP module (\#5) produces opposite outcomes, particularly with respect to the SR metric. This may be attributed to the learnable token $v$, which, while pooling hybrid environmental representations, also introduces slight noise that causes fluctuations of navigation performance. Nevertheless, the overall performance is progressively improved with the inclusion of DSP, TSU and HAP modules.


\subsection{Qualitative Analysis }
\begin{figure}
    \centering
    \includegraphics[width=0.98\linewidth]{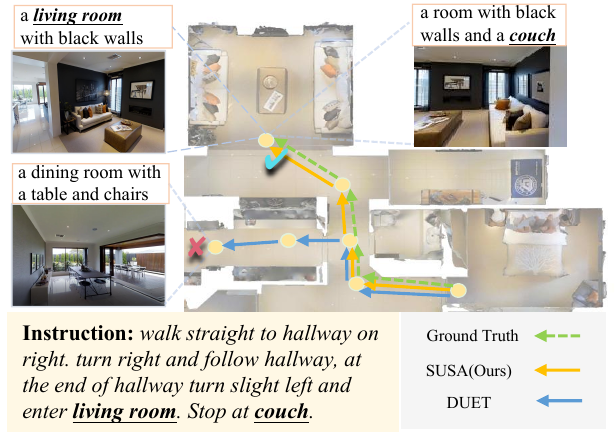}
    \caption{Predicted trajectories of SUSA and DUET on the R2R validation unseen split.}
    \label{fig:vis}
\end{figure}

As shown in Fig.~\ref{fig:vis},  compared to DUET, our SUSA agent is able to accurately stop near the ``living room with a couch." In contrast, the DUET agent, which relies exclusively on RGB environment representations, is prone to making elusive actions, hindering its ability to accurately follow instructions and reach the target location. This manifests that, thanks to the salient landmarks provided by the SUSA architecture, the agent is better able to navigate to the target location by grounding textual instructions.


\section{Conclusion}
We present the SUSA architecture, which hierarchical exploits semantic-spatial representations beyond RGB to facilitate environmental understanding and instruction grounding. Specifically, we propose the TSU module, enabling the agent to identify the most instruction-relevant textual semantic. The DSP module complementarily enhances spatial awareness by incrementally constructing and grounding depth exploration maps.
Experimental results on the R2R, REVERIE, SOON and R2R-CE benchmarks demonstrate the navigational superiority of our approach.
Benefiting from hierarchical representation enrichment and alignment, SUSA exhibits excellent generalization performance on unseen environments and modality interpretability.

\section{Acknowledgments}
This work was supported by the National Natural Science Foundation of China (NSFC) under Grants Nos. 62202139, 62573292 and 62172138, and also supported by the Fundamental Research Funds for the Central Universities under Grants Nos.JZ2025HGTB0226, JZ2024HGTG0310 and Young Elite Scientist Sponsorship Program (No. YESS20220409). The computations were performed on the High-Performance Computing (HPC) Platform of Hefei University of Technology.

\bibliography{aaai2026}

\appendix

\section*{Appendices}
\label{sec:suppl}

This appendix provides additional materials to complement the main paper. Specifically:
\begin{itemize}
    \item Section~\ref{A} presents detailed descriptions of metrics, dataset details and pre-training tasks.
    \item Section~\ref{B} offers extended experimental results, including: (1) comparisons with related approaches, (2) analysis of the learned fusion weights across modalities, (3) evaluation of the quality of generated salient landmarks, (4) ablations on the contrastive loss weight, (5) an overview of the overall design on the R2R benchmark, and (6) a discussion on the trade-off between computational efficiency and navigation performance.
    \item Section~\ref{C} showcases additional qualitative examples to analyse the agent’s behavior.
    \item Section~\ref{D} discusses the limitations of our current approach and outlines potential directions for future work.
\end{itemize}


\section{Detailed Experimental Setups}\label{A}

\noindent \textbf{Metrics.}
We categorize the metrics into three types: \textit{Navigation}, \textit{Objects Grounding}, and \textit{Instruction Following}. Detailed calculations of these metrics can be found in Table~\ref{tab:metric}.
For \textit{Navigation} metrics, Trajectory Length (TL): Average path length in meters. It is worth noting that since the trajectory needs to match the ground truth trajectory described by the instruction, strictly speaking, the trajectory length is not necessarily the smaller, the better. 
Success Rate (SR): Proportion of paths where the agent reaches within 3 meters of the target location.
Oracle Success Rate (OSR): The proportion of trajectories where at least one node is within 3 meters of a corresponding node in the reference trajectory.
Navigation Error (NE): Average final distance (meters) between the agent and the target.
Success Rate weighted by Path Length (SPL): Prioritizes success for shorter paths, which normalizes the success rate by trajectory length.
For \textit{Instruction Following} metrics, we apply Normalized Dynamic Time Warping (nDTW) and Success Rate Weighted by Dynamic Time Warping (sDTW). The former measures how well the VLN agent could follow the instruction, 
while the latter considers nDTW for success cases.
For \textit{Objects Grounding} metrics: Remote Grounding Success (RGS): Proportion of instructions where the agent correctly identifies the target object.
RGS weighted by Path Length (RGSPL): Prioritizes correct object identification for shorter paths. 

\begin{table}[ht]
    \centering
    \begin{tabular}{c|cc}
    \toprule
    \textbf{Metric}   & \textbf{$\uparrow$ $\downarrow$} & \textbf{Definition} \\ \midrule
    \multicolumn{3}{l}{\ \ \ \ \ \  \textit{Navigation:}}  \\ \midrule
    
    TL   & & $\sum_{1\leq i<|P|}d(p_i,p_{i+1})$  \\
    NE  &  $\downarrow$   &  $d(p_{|P|},g_{|G|})$   \\
    ONE &  $\downarrow$   & $\min_{p\in P}d(p,g_{|G|})$    \\
    SR   &  $\uparrow$   & $\mathbb{I}[NE(P,G) \leq d_{th}]$   \\
    OSR &    $\uparrow$ & $ \mathbb{I}[\mathrm{ONE}(P,G)\leq d_{lh}]$    \\
    SPL &  $\uparrow$   &  $\mathrm{SR}(P,G)\cdot\frac{d(p_1,g_{|G|})}{\max\{\mathrm{TL}(P),d(p_1,g_{|G|})\}}$   \\ \midrule
    \multicolumn{3}{l}{\ \ \ \ \ \  \textit{Instruction Following:}}  \\ \midrule
    nDTW &  $\uparrow$   & $\exp\left(-\frac{\min_{W\in\mathcal{W}}\sum_{(i_{k},j_{k})\in W}d(p_{i_{k}},q_{j_{k}})}{|P|\cdot d_{th}}\right)$   \\
    sDTW &  $\uparrow$   &  $SR(P,G)*nDTW(P, G)$   \\ \midrule
    \multicolumn{3}{l}{\ \ \ \ \ \  \textit{Objects Grounding:}}  \\ \midrule
    RGS &   $\uparrow$  &  $\mathbb{I}\Big(\max_jIoU(B_i^p,B_j^g)\geq0.5\Big)$   \\
    RGSPL &  $\uparrow$   & $\mathrm{RGS}(P,G)\cdot\frac{d(p_1,g_{|G|})}{\max\{\mathrm{TL}(P),d(p_1,g_{|G|})\}}$   \\ \bottomrule
    \end{tabular}
    \caption{Common VLN metrics, categorized into three types: Navigation, Instruction Following, and Object Grounding. The symbols $\uparrow$ and $\downarrow$ signify if a higher or lower metric value is preferable, respectively. P represents the paths taken by the agent, while G stands for the ground truth trajectories. The function $\mathbb{I}(\cdot)$  serves as an indicator, with $d(n, P)$ measuring the distance from node $n$ to path $P$ , and $d_{th}$ indicating the distance threshold. $W=w_{1..|W|}$ is a warping with $w_k=(i_k,j_k)\in[1:|P|], [1:|G|]$, respecting the step-size.$B^p_i$ means the i-th object's bounding box in agent path (or trajectory).}
    \label{tab:metric}
\end{table}

\begin{table*}[htb]
\centering
\begin{tabular}{c|cccccccc|c}
\toprule
\multirow{2}{*}{\textbf{Dataset}} & \multicolumn{2}{c}{\textit{\textbf{Train}}} & \multicolumn{2}{c}{\textit{\textbf{Val Seen}}} & \multicolumn{2}{c}{\textit{\textbf{Val Unseen}}} & \multicolumn{2}{c|}{\textit{\textbf{Test Unseen}}} & \multirow{2}{*}{\textbf{VLN task}} \\ 
                         & instr            & house           & instr                 & house                & instr                  & house                 & instr               & house               &                           \\ \midrule
R2R  \cite{anderson2018R2R}                    & 14039            & 61              & 1021                  & 56                   & 2349                   & 11                    & 4173                & 18                  & Fine-grained              \\
R2R-CE \cite{krantz_vlnce_2020}                    & 10819            &  61            & 778                   & 53                   & 1839                   & 11                   & 3408               & 18                 & Fine-grained           \\ 
REVERIE \cite{qi2020reverie}                 & 10466            & 60              & 1423                  & 46                   & 3521                   & 10                    & 6292                & 16                  & Goal-oriented             \\
SOON \cite{zhu2021soon}                    & 2779             & 34              & 113                   & 2                    & 339                    & 5                     & 615                 & 14                  & Goal-oriented            \\ \bottomrule
\end{tabular}
\caption{Comparison of the three VLN datasets utilized in this study.}
    \label{tab:datasets}
\end{table*}


\noindent \textbf{Dataset Details.}
We primarily evaluate the agent's navigation capabilities on three distinct VLN datasets: the fine-grained R2R~\cite{anderson2018R2R}, and the goal-oriented REVERIE~\cite{qi2020reverie} and SOON~\cite{zhu2021soon}. To further assess generalization in continuous environments  \cite{savva2019habitat}, we extend our evaluation to the R2R-CE~\cite{krantz_vlnce_2020} task, as shown in Table~\ref{tab:datasets}.
Each dataset is divided into train, validation seen, validation unseen, and test unseen splits. REVERIE contains approximately 20k high-level instructions describing target locations and objects, with an average instruction length of 21 words. Given predefined object bounding boxes within each panorama, the agent identifies the correct bounding box upon reaching the navigation goal. Unlike the R2R dataset, in the REVERIE task, the agent must identify the correct bounding box from the candidates once the navigation target is reached. R2R includes 22k step-by-step instructions, with an average instruction length of 32 words. The R2R dataset focuses solely on the navigation task, providing detailed path planning for the agent in the instructions. In R2R-CE, the average distance between navigable nodes is significantly shorter compared to R2R (0.19 vs. 2.25 meters), while the number of navigation steps is substantially higher (55 vs. 4–6). SOON is another goal-oriented dataset, containing about 5k instructions, each averaging 47 words in length.

\noindent{\textbf{Pre-training Tasks.}}
Following prior works~\cite{chen2022duet,chen2021hamt}, during pre-training, we employ a suite of multimodal pre-training tasks, including MLM (Masked Language Modeling), MRC (Masked Region Classification), SAP (Single Action Prediction), and OG (Object Grounding). OG is specifically designed for object grounding using the REVERIE and SOON datasets. MLM and MRC aim to establish a robust vision-language alignment. SAP enables the agent to learn preliminary navigation knowledge by predicting the next action during pre-training. However, as navigation is inherently a sequential decision-making process, SAP can only learn static, single-step decisions. When transferring to the R2R-CE task, the agent—serving as part of the planner module—is not trained from scratch in continuous environments, but only fine-tuned.

We evaluated the experimental results of the unseen test split on the public online leaderboards\footnote{\url{https://eval.ai/web/challenges/challenge-page/606/participate}, \url{https://eval.ai/web/challenges/challenge-page/97/participate}.}.
We assure the readers that we will make our code and model checkpoints publicly available.

\section{Additional Experiments}
\label{B}

\begin{table}[]
\centering
\setlength\tabcolsep{5pt}
\begin{tabular}{cc|cccc}
\toprule
\multicolumn{2}{c|}{}                                                              & \multicolumn{4}{c}{\textbf{Val Unseen}}                                       \\
\multicolumn{2}{c|}{\multirow{-2}{*}{\textbf{Methods}}}                                                                                       & SR   & \textbf{SPL} & RGS  & RGSPL                           \\ \midrule
 & BEVBert  & 51.78                                                & 36.37                                                & 34.71                                                & 24.44                                                \\
   & GridMM                                               & 51.37                                                & 36.47                                                & 34.57                                                & 24.56                                                \\
 & SEAT & 49.45& 35,51& 32.83&23.14\\
  &  \textbf{SUSA(DP)}   & \textbf{52.54} & 37.59 & \textbf{35.47} & 25.27 \\
\multirow{-4}{*}{Depth} &  \textbf{SUSA(DEM)}  &  52.00 & \textbf{39.21} & 35.05 &  \textbf{26.54} \\ \midrule
  & KESU                                             & 50.36                                                & 35.15                                                & 35.02                                                & 24.17                                                \\
  & KERM                                                & 50.44                                                & 35.38                                                & 34.51                                                & 24.45                                                \\
 & AACL &49.42 &33.54 &33.31 &22.49\\
  & CONSOLE                                                   & 50.07                                                & 34.40                                                & 34.05                                                & 23.33                                                \\
\multirow{-4}{*}{Text}  & \textbf{SUSA(TP)}                                               & \textbf{53.17}& \textbf{38.31}& \textbf{37.18}& \textbf{26.73}\\ \bottomrule
\end{tabular}
\caption{Comparison with related works on the REVERIE validation unseen split.}
\label{tab:related}
\end{table}


\subsection{Comparison with Related Approaches}

We demonstrate the effectiveness of the proposed method through fair comparisons with related approaches. As shown in Table~\ref{tab:related}, experiments reveal that directly grounding depth information—whether using depth panoramas (DP) or depth exploration maps (DEM) as input—outperforms prior depth-based methods \cite{an2023bevbert,wang2023gridmm,wang2024enhanced}. Notably, using depth exploration maps leads to better navigation performance than depth panoramas. Similarly, directly aligning textual semantics with instructions via textual panoramas (TP) surpasses methods \cite{gao2024enhancing,li2023kerm,lin2024correctable} that integrate textual semantics into environmental representations. This underscores the superiority of independently grounding each modality, improving the interpretability of each modality’s contribution.

\begin{figure}[ht]
    \centering
    \includegraphics[width=1\linewidth]{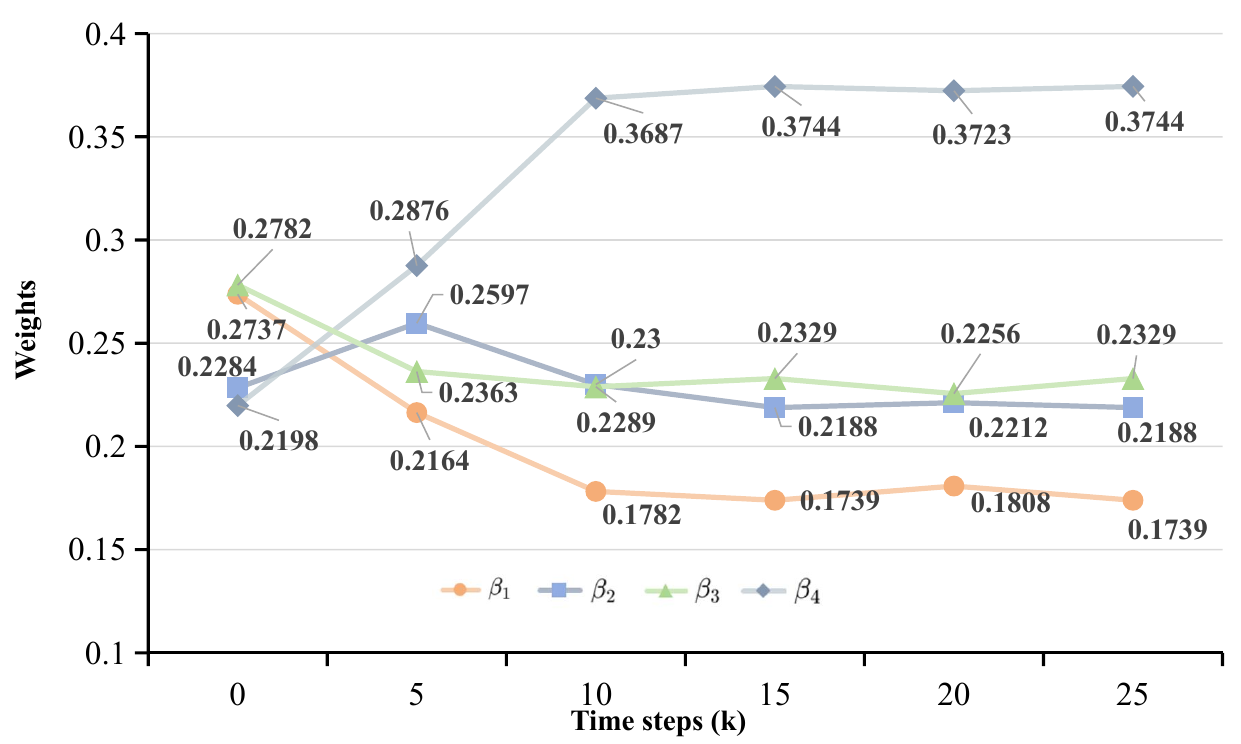}
    \caption{The weights for hybrid representations in the HAP module change during training (see Eq. ~\ref{eq: branch-weights}).}
    \label{fig:weight}
\end{figure}

\begin{figure*}[h]
    \centering
    \includegraphics[width=1\linewidth]{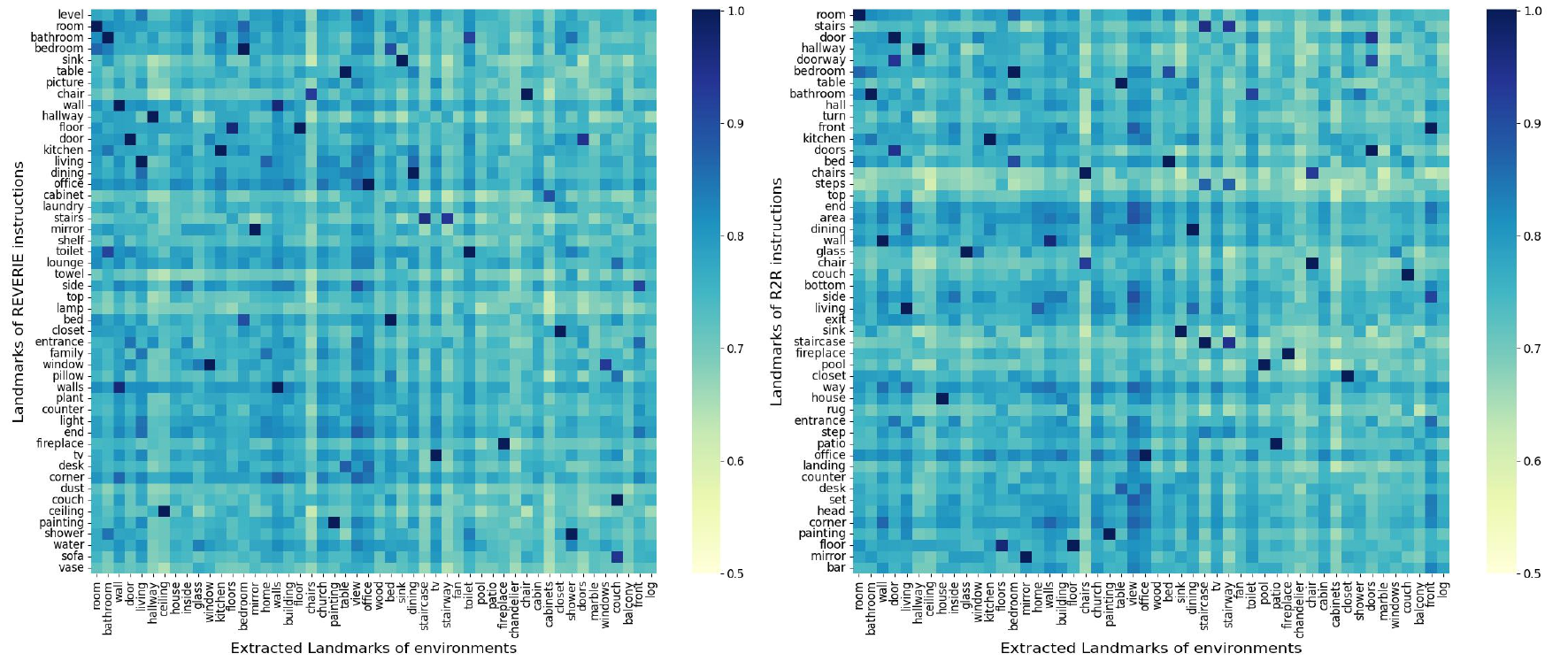}
    \caption{Similarity matrix for the top 50 most frequent landmarks: REVERIE instructions vs. environments (left), R2R instructions vs. environments (right). Best viewed in color. }
    \label{fig:landmark}
\end{figure*}
\subsection{Fusion Weights of Hybrid Representations.}
We quantitatively illustrate the weights (see Eq.~\ref{eq: branch-weights}) assigned to each branch of the HAP module during training (in Fig.~\ref{fig:weight}) and inference (in Table~\ref{tab:weight}). These weights are normalized between 0 and 1, with higher values denoting stronger dependency on the corresponding environmental representation. 
As shown in Fig.~\ref{fig:weight}, initially, all four representations have approximately equal weights of 0.25 during training, reflecting random initialization and balanced contributions. Over the course of training, the global RGB representation weight $\beta_4$ increases, while those of the textual semantic $\beta_1$ and depth $\beta_3$ modalities decrease. This trend arises because textual semantics and depth provide sparse yet precise information, which is particularly beneficial in the early stages of training. Despite the increasing dominance of the RGB modality, the agent's reliance on textual semantics and depth remains above 20\%, underscoring their essential role in navigation. Table \ref{tab:weight} also demonstrates that the agent effectively utilizes the introduced depth spatial and textual semantic information during inference, even though it exhibits a stronger reliance on the RGB modality.
\begin{table}[t]
\centering
\begin{tabular}{c|cccc}
\toprule
\multirow{2}{*}{\textbf{Splits}} & \multicolumn{2}{c}{\textbf{Local}} & \multicolumn{2}{c}{\textbf{Global}} \\
                        & $\beta_1$& $\beta_2$ & $\beta_3$ &$\beta_4$ \\ \midrule
Val Seen         & 0.2147      & 0.2458    & 0.2239& 0.3156\\
Val Unseen       & 0.2151     & 0.2662    & 0.2216& 0.2971\\ \bottomrule
\end{tabular}
\caption{Variation in fusion weights (in Eq.~\ref{eq: branch-weights}) of hybrid representations during inference on R2R. $\beta_1$, $\beta_2$, $\beta_3$, $\beta_4$ denotes the fusion weight of textual semantic, local RGB, depth, global RGB representations in the HAP module, respectively. }
\label{tab:weight}

\end{table}

\begin{figure}[t]
    \centering
    \includegraphics[width=1\linewidth]{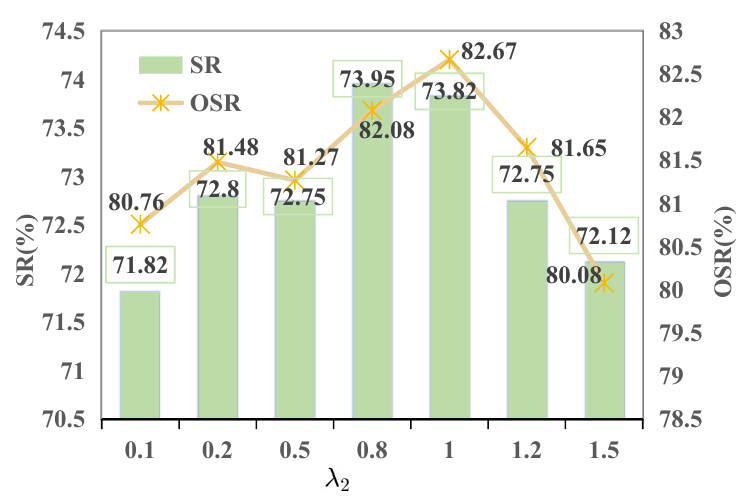}
    \caption{Analysis of the contrastive loss weight $\lambda_2$ in HAP module on R2R validation unseen split.}
    \label{fig:loss-weight}
\end{figure}

\subsection{Quality of Generated Salient Landmarks}
The quality of the extracted salient landmarks is crucial for the agent to accurately understand the textual semantics of the environment.
In Fig.~\ref{fig:landmark},
we compute the similarity between the top-50 most frequently occurring landmarks in our generated salient landmarks and those from R2R \cite{anderson2018R2R} and REVERIE \cite{qi2020reverie} to evaluate the quality of the generated salient landmarks.
While the types and frequencies of these generated landmarks is different from those in the datasets, they can contain and convey most similar sematic (similarities surpass 0.5). 
Notably, the plug-and-play CLIP text encoder excels in understanding image-related descriptions, while BERT is more adept at comprehending natural language sentences. Therefore, we use the CLIP text encoder to extract features from the generated salient landmarks. 

\begin{table}[t]
\centering
\setlength\tabcolsep{3pt}
\begin{tabular}{c|ccc|cccc}
\toprule
\ ID \ &\ DSP & \ TSU & HAP \ & \ \ TL   & \ OSR$\uparrow$  & \ SR$\uparrow$   & \ \textbf{SPL}$\uparrow$ \ \ \\ \midrule
1  & \ding{55}                        & \ding{55}                         & \ding{55}   & 13.41& 79.86 & 71.43& 60.63\\
2  & \ding{51} & \ding{55} & \ding{55}   & 13.85 & 80.42 & 72.80 & 62.71       \\
3  & \ding{55} & \ding{51} & \ding{55}   & 13.78 & 81.65 & 72.92 & 63.16 \\
4  & \ding{51} & \ding{51} & \ding{55}   & 12.18 & 79.99 & 73.05 & \textbf{64.85} \\
5  & \ding{51} & \ding{51} & \ding{51} & 14.58 & \textbf{82.08} & \textbf{73.95} & 62.84 \\ \bottomrule
\end{tabular}
\caption{Ablation study on different components of SUSA on the R2R validation unseen split.}
\label{tab:overall-r2r}
\end{table}

\subsection{Analysis of the Contrastive Loss Weight}
Our ablation studies indicate that the contrastive learning loss weight, denoted as $\lambda_2$ in Eq.~\ref{loss}, affects model navigation performance.  As shown in Fig.~\ref{fig:loss-weight}, agent achieves the optimal navigation performance when the $\lambda_2$ is set to 0.8. we further observe that, within a certain range, increasing the contrastive loss weight improves navigation performance. However, when the contrastive loss weight is too high, performance (SR/OSR) declines, thus overemphasizing contrastive learning loss may lead the agent to prioritize modality alignment over core sequential navigation tasks.

\subsection{Overall Design on R2R}
We further supplemented the performance of each key component of SUSA on the R2R task. 
As shown in Table~\ref{tab:overall-r2r}, our previous conclusions are reaffirmed: the DSP module allows the agent to better perceive spatial layouts, thus improving navigation efficiency (SPL). The TSU module is better at understanding textual semantics, resulting in higher success rates.
After integrating the DSP and TSU modules, the agent's performance improved significantly (e.g., SR=73.05, SPL=64.85). 
When the HAP module is further introduced, the navigation success rate on R2R increases to 73.95, indicating that contrastive learning further enhances instruction-environment alignment.
However, its SPL metric remains suboptimal, suggesting that the contrastive learning strategy in the HAP module may excel at accurately aligning the environment with instructions but lacks the capability to optimize the navigation path length.

\subsection{Computing Efficiency vs. Navigation Performance}
\begin{figure}
    \centering
    \includegraphics[width=1\linewidth]{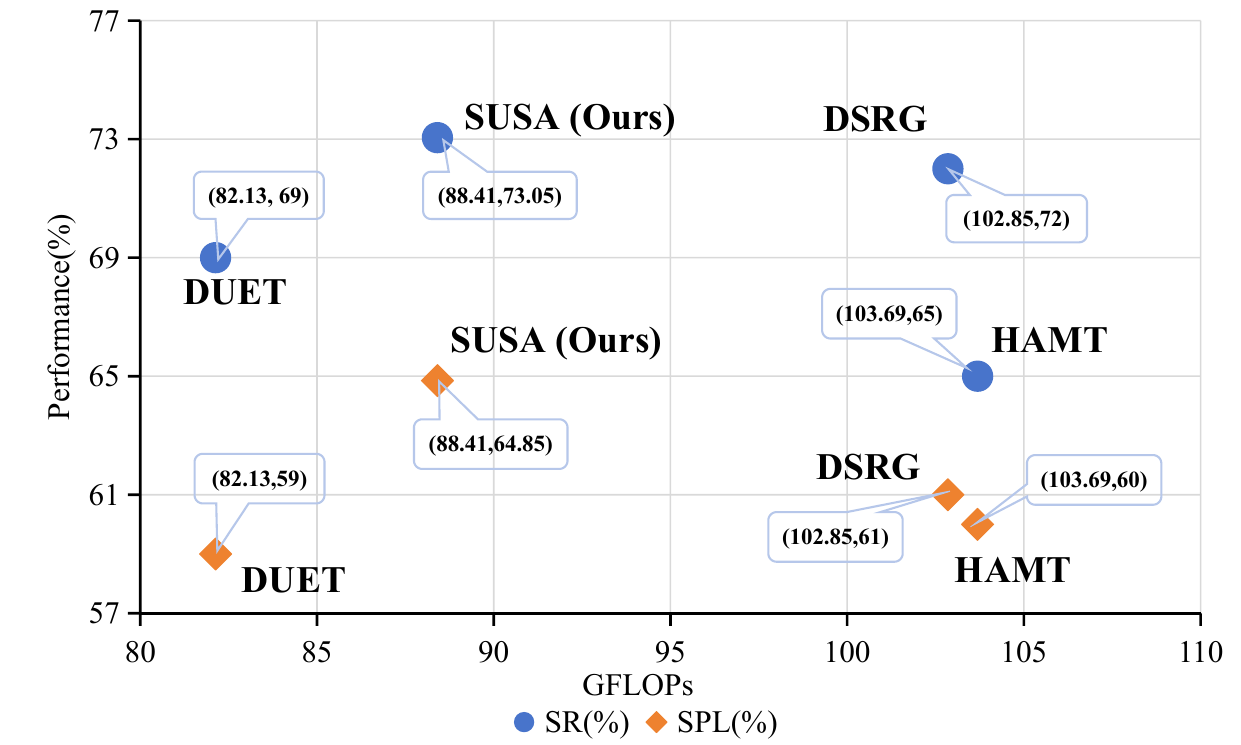}
    \caption{Comparison of computational efficiency and navigation performance on R2R validation unseen split with several popular VLN methods.}
    \label{fig:GFLOPs}
\end{figure}
As~\ref{fig:GFLOPs} illustrated, we compare the Giga Floating-point Operations (GFLOPs) and navigation performance (i.e, SR/SPL, which are more important metrics) of different VLN models, including DUET \cite{chen2022duet} , DSRG \cite{wang2023dsrg},  HAMT \cite{chen2021hamt} and our SUSA. To ensure a fair comparison across all methods, we performed single-step forward inference with a batch size of 8, an instruction length of 44, and 6 exploration map nodes.
Compared to the baseline DUET, our approach introduces spatial and semantic environmental information, which inevitably increasing computational cost. However, benefiting from the partial pretraining strategy and a shared panorama encoder for both the depth and RGB environmental information, our SUSA architecture notably enhances agent navigation performance with modest computational resources.
The frozen BLIP-2 and CLIP models provide the SUSA agent with off-the-shelf textual semantics and are excluded from the training or inference.



\section{Qualitative Examples} 
\label{C}

\begin{figure}
    \centering
    \includegraphics[width=0.95\linewidth]{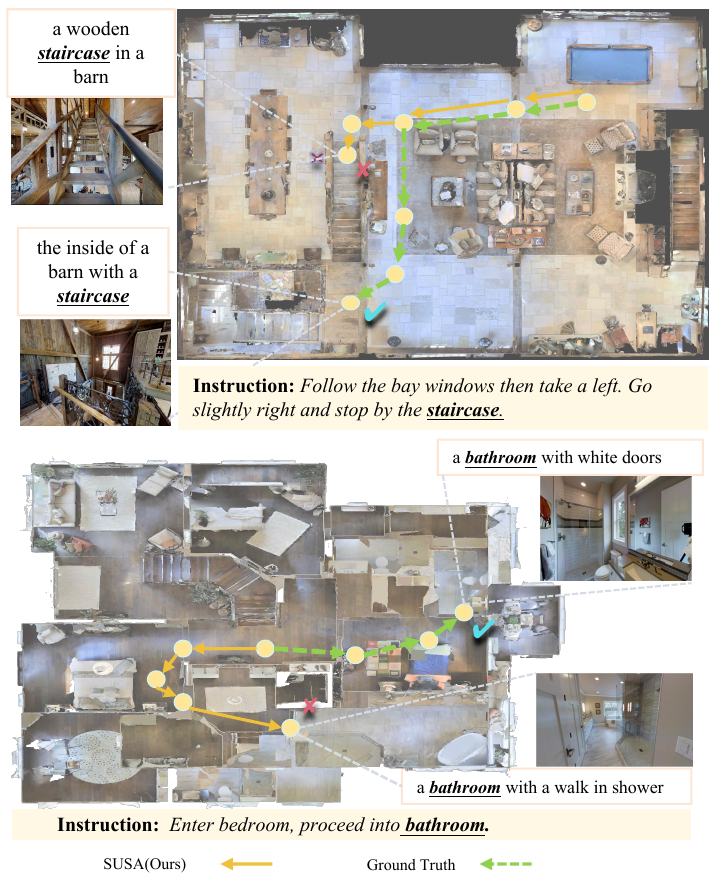}
    \caption{Top-down trajectories visualization of failed cases. Partial views of final stop position and corresponding landmark captions are also provided.}
    \label{fig:failure case}
\end{figure}

\noindent \textbf{Failure cases.}
We provide an in-depth examination of failure cases, aiming to offer valuable insights for future model improvements. \ref{fig:failure case} displays the difference between the navigation trajectory predicted by the proposed SUSA and the ground truth trajectory. We found that, although textual semantics can provide the agent with a richer understanding, it may introduces ambiguity in certain scenarios. Concretely, when two identical salient landmarks appear in different scenes (e.g., the staircase and bathroom in~\ref{fig:failure case}), the agent may struggle to distinguish the correct navigable node, leading to erroneous navigation decisions.

\begin{figure*}[t!]

    \captionsetup{justification=justified, singlelinecheck=false}

    \includegraphics[width=\linewidth]{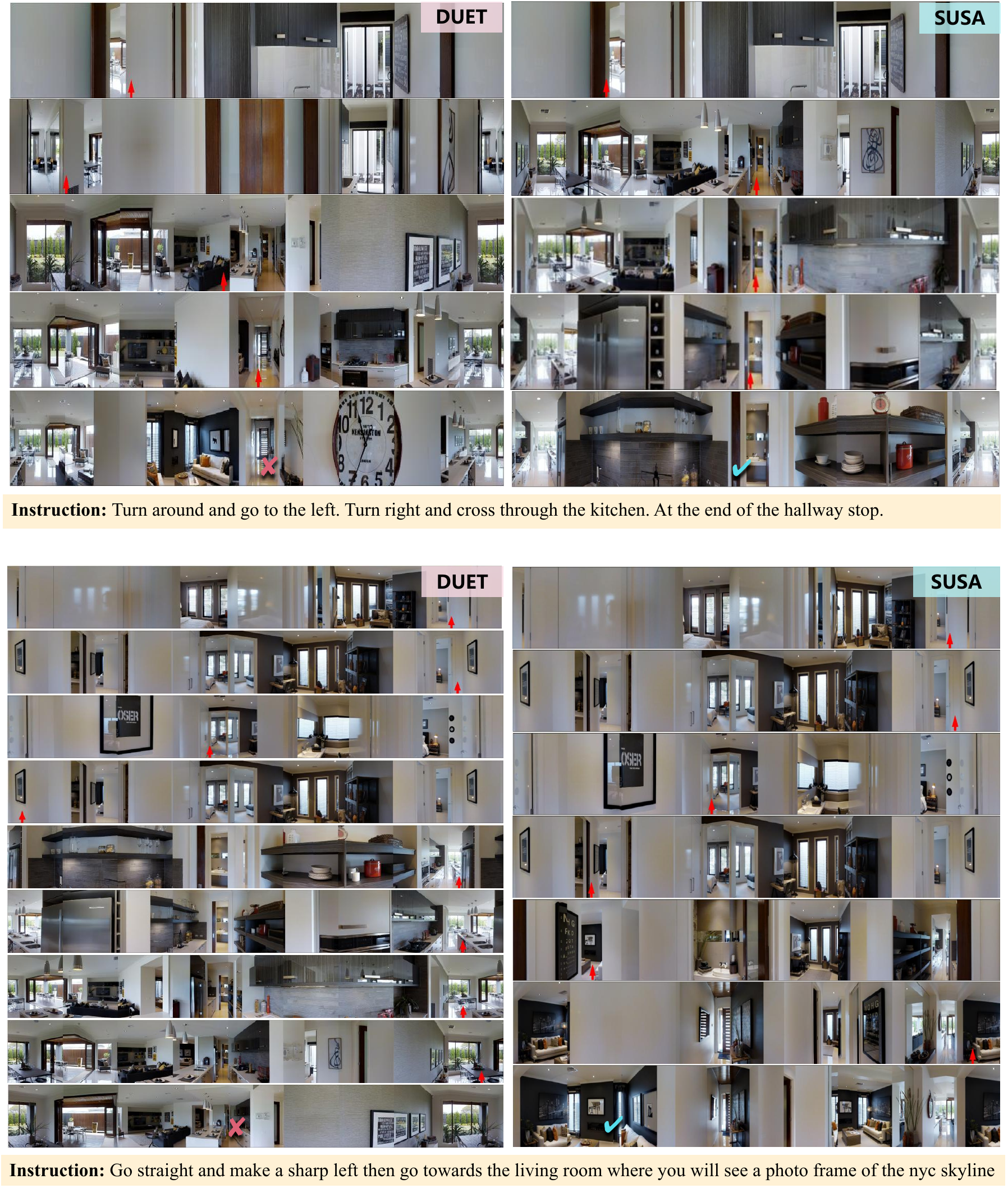}
    \caption{First-person trajectory visualizations of DUET (left) and our SUSA (right), with the corresponding instructions provided below. Red arrows indicate the next selected viewpoint from current position.}
    \label{fig:first-1}
\end{figure*}

\noindent \textbf{More Qualitative Examples.}
We further visualize first-person navigation trajectories on the R2R validation unseen split, as illustrated in Fig.~\ref{fig:first-1}.





\section{Limitations and Future Work}
\label{D}

As previously mentioned, when an agent encounters the same landmark at different navigable points, it may struggle to make confident navigation decisions, leading to failures. To address this limitation, 1) our future research will explore adjusting the confidence levels associated with different environmental representations to help the agent make more accurate and reliable predictions. 2) Beyond landmarks, we will construct direction-related textual semantics (e.g., ``past," ``turn left") to better model the spatial relationships between actions in instructions and the environmental layout or historical exploration maps. 

In the future, we also intend to explore the generalizability of our approach, which hierarchically  integrates richer representations from multiple perspectives, in embodied tasks such as continuous navigation and embodied question answering.
Additionally, given the complexity of VLN tasks, current mainstream VLN frameworks are often intricate and cumbersome, with a wide variety of evaluation metrics that 
are difficult to balance.
Ongoing research in the future is expected to develop more lightweight VLN model architectures and comprehensive evaluation metrics.
\clearpage

\clearpage

\end{document}